\documentclass[letterpaper]{article}
\usepackage{aaai2027}
\usepackage[hyphens]{url}
\usepackage{graphicx}
\usepackage{natbib}
\usepackage{amsmath}
\usepackage{amssymb}
\usepackage{booktabs}
\usepackage{caption}
\usepackage{xcolor}
\usepackage{colortbl}
\usepackage{tabularx}
\usepackage{microtype}
\newcolumntype{Y}{>{\centering\arraybackslash}X}
\copyrighttext{\textsuperscript{*}Equal contribution. \textsuperscript{\textdagger}Project leader. \textsuperscript{\textdaggerdbl}Corresponding author.}
\title{DriveCache: Action-Aware Caching for Driving World Model Inference}
\author{
Jianchun Yang\textsuperscript{\rm 1,*},
Jian Liang\textsuperscript{\rm 1,*},
Xianda Guo\textsuperscript{\rm 1,*,\textdagger},
Pinhan Fu\textsuperscript{\rm 1},
Yanlun Peng\textsuperscript{\rm 3},
Conglang Zhang\textsuperscript{\rm 1}, \\
Wenke Huang\textsuperscript{\rm 2},
Mang Ye\textsuperscript{\rm 1,\textdaggerdbl}
}
\affiliations{
\textsuperscript{\rm 1}Wuhan University \quad
\textsuperscript{\rm 2}Nanyang Technological University \quad
\textsuperscript{\rm 3}Great Wall Motor \\
yangjianchun@whu.edu.cn, jianliang@whu.edu.cn, xianda\_guo@163.com, yemang@whu.edu.cn
}

\begin{document}
\maketitle
\setlength{\abovedisplayskip}{5pt plus 2pt minus 4pt}
\setlength{\belowdisplayskip}{5pt plus 2pt minus 4pt}
\setlength{\abovedisplayshortskip}{0pt plus 2pt}
\setlength{\belowdisplayshortskip}{3pt plus 2pt minus 2pt}

\begin{abstract}
Driving video generation models support autonomous-driving development by predicting controllable future scenes for simulation, planning evaluation, and offline data generation. Diffusion-based driving generators repeatedly evaluate large backbones across denoising steps, which limits generation throughput. Existing diffusion acceleration methods reduce this cost, but general-purpose designs omit driving signals available before generation, such as ego speed and planned trajectories. Experiments across driving motions show that cache tolerance varies with ego translation and rotation, denoising progress, and consecutive reuse length. We propose DriveCache, a training-free, action-aware controller that uses planned motion to allocate reuse across scenes and dynamic programming to place it across denoising steps under a calibrated response budget. A causal drift check refreshes features and replans the remaining schedule when generation departs from calibration. Across three generator configurations, DriveCache improves the overall fidelity--efficiency trade-off over evaluated cache methods. \mbox{Our code will be publicly available.}
\end{abstract}
\section{Introduction}
Driving video generation models predict how traffic scenes evolve under different ego actions, providing controllable future observations for simulation, policy training, planning evaluation, and offline data generation \citep{hu2023gaia,wang2024drivedreamer,gao2024vista}. These action-conditioned predictions let developers and planning systems examine possible futures under candidate plans. They also support diverse scenario generation for development. Recent video diffusion models improve visual fidelity and temporal coherence through larger spatiotemporal backbones and longer generation horizons \citep{ho2022video,blattmann2023stable,yang2025cogvideox,kong2024hunyuanvideo,lin2024open}.

These advances increase inference cost. Iterative denoising evaluates the backbone across many sampling steps, and the cost grows with model capacity, resolution, and video length. Online prediction must fit within the planning cycle, while offline generation must scale to large scenario collections. Inference latency therefore constrains online prediction and offline scenario generation.

Diffusion acceleration methods reduce the number or cost of denoising evaluations. Distillation, quantization, pruning, and efficient operators can require new weights, retraining, calibration data, or specialized kernels \citep{salimans2022progressive,song2023consistency}. Feature caching leaves the generator unchanged and reuses intermediate features across neighboring denoising steps. Existing cache controllers derive reuse from fixed schedules or signals observed after denoising begins \citep{liu2025timestep,kahatapitiya2025adaptive,zhou2025easycache,ma2026magcache,bu2025dicache,liu2025reusing,chung2026seacache}. These general-purpose methods omit driving signals available before generation, including ego speed and planned trajectories.
\begin{figure}[t]
\centering
\includegraphics[width=\columnwidth]{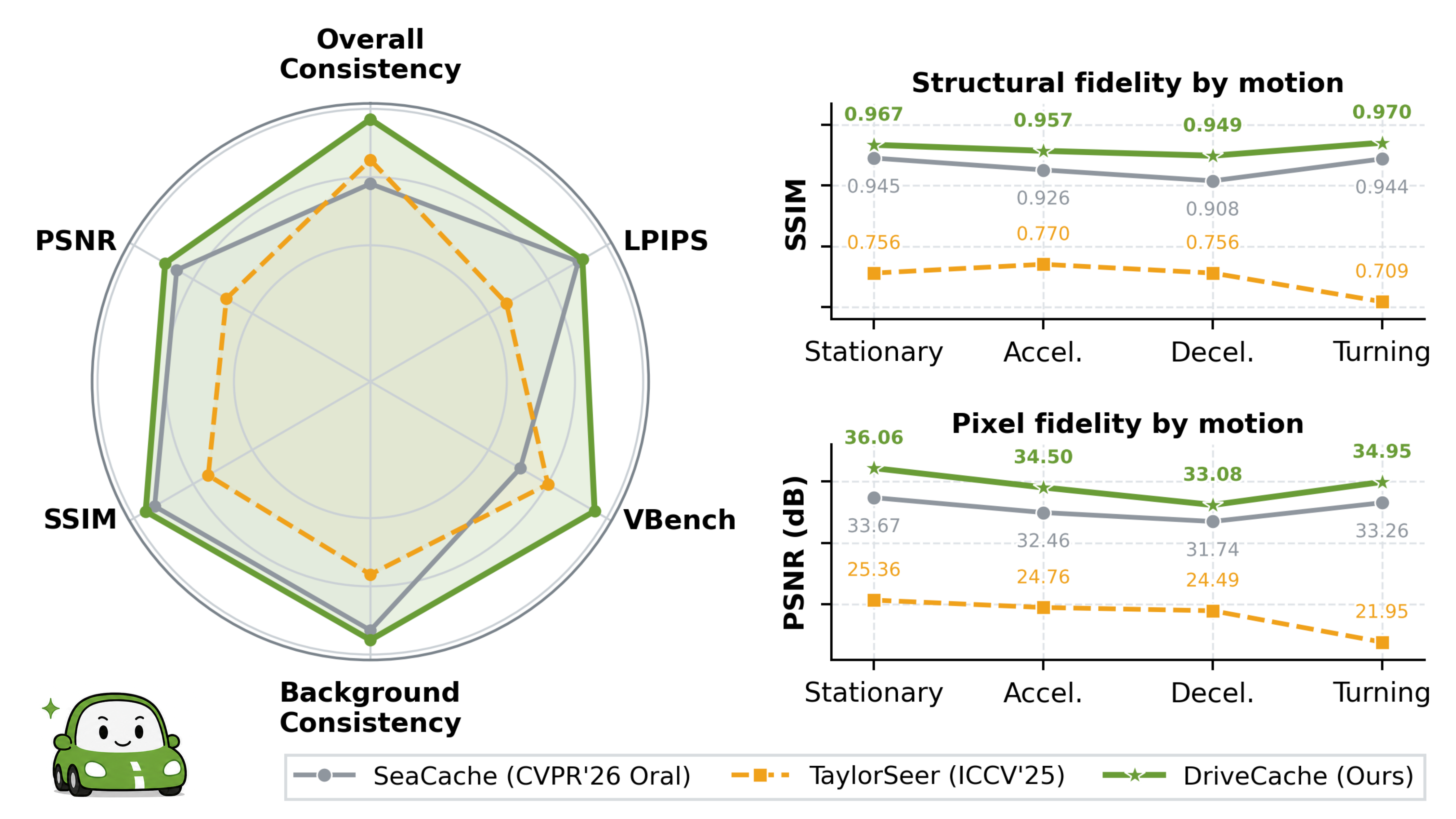}
\caption{DriveCache achieves better video fidelity and consistency across evaluation metrics.}
\label{fig:teaser}
\end{figure}
\begin{figure*}[t]
\centering
\includegraphics[width=0.99\textwidth]{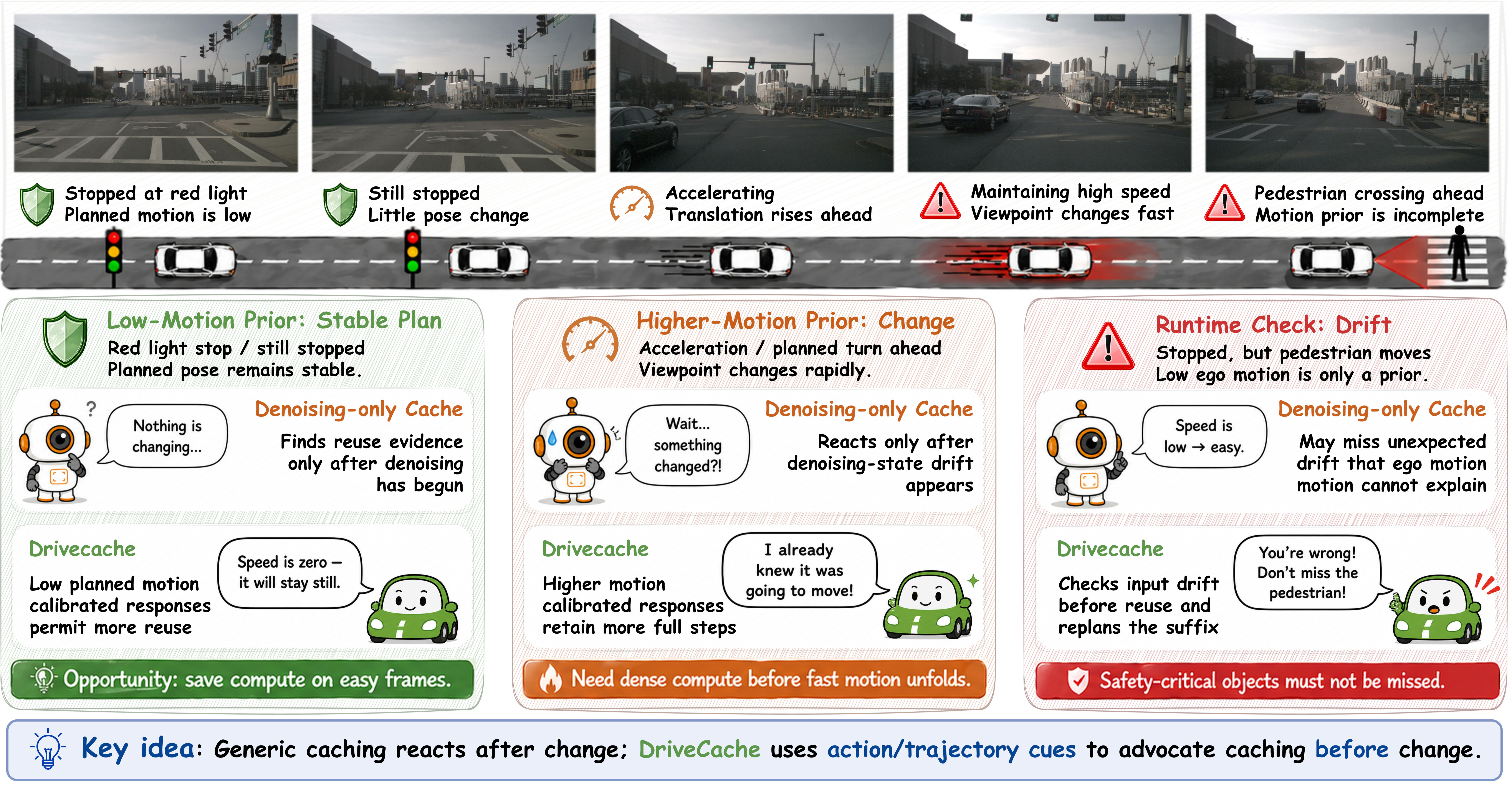}
\caption{Motivation of DriveCache. Planned ego motion reveals scene-dependent cache tolerance before denoising.}
\label{fig:motivation-overview}
\end{figure*}
However, driving generation provides planned ego motion before denoising. Offline generation receives recorded or specified future ego poses, while planner-conditioned generation receives predicted poses from the planning stack. Stationary, straight, and turning plans induce different viewpoint changes. Figure~\ref{fig:motivation-overview} shows that planned ego motion predicts scene-dependent cache tolerance before the denoising pass begins.

Planned motion is available before the first denoising step, but it cannot determine a schedule by itself. Cache error also varies with denoising position, cache age, model architecture, and generated content. A driving-aware controller must therefore combine a scene-level motion prior with a denoising-level response model, while retaining a causal correction for states that depart from calibration.

DriveCache converts this prior into a complete schedule. One low-motion anchor and one moving-turn anchor measure terminal responses of consecutive reuse; planned translation and rotation interpolate them for each scene. Exact dynamic programming selects reuse quantity and positions under one budget. A pre-reuse drift veto rejects out-of-support decisions and replans the unexecuted suffix.

Our main contributions are threefold.
\begin{itemize}
\item To the best of our knowledge, this work is the first to identify and validate planned ego motion as a pre-generation signal for diffusion caching in driving video generation, showing that ego translation and rotation predict scene-level cache tolerance across driving motions.
\item We propose DriveCache, a training-free cache controller that assigns scene-level reuse budgets from planned motion, models consecutive-reuse error, and uses exact dynamic programming with causal drift correction to place reuse across denoising steps.
\item Across multiple driving video generators, DriveCache improves quality--efficiency trade-offs over cache baselines. At approximately \(2\times\) speedup on Wan2.2 A14B, it improves PSNR by 2.036 dB over TeaCache.
\end{itemize}

\section{Related Work}
\subsection{Diffusion Models}

Diffusion models learn a reverse transport from noise to data \citep{ho2020denoising,song2021scorebased,rombach2022high}. Deterministic samplers and continuous-time formulations broaden this process \citep{song2020denoising,Karras2022edm,lu2022dpm,lipman2022flow}. Video diffusion adds spatiotemporal modeling \citep{ho2022video,blattmann2023stable,ma2024latte,yang2025cogvideox,kong2024hunyuanvideo,lin2024open}, while Diffusion Transformers scale generation \citep{peebles2023scalable,chen2024pixart}. Token merging and low-precision attention lower per-step cost \citep{bolya2022token,zhang2024sageattention}. Training-based methods distill samplers or generation dynamics but require optimized weights and model-specific training \citep{salimans2022progressive,song2023consistency,luo2023latent,yin2024improved,yin2025causvid}. These acceleration methods change sampling, reduce token or attention cost, or train modified generator weights.

\subsection{Training-free Diffusion Inference Acceleration}

Training-free acceleration preserves pretrained weights. Fast solvers reduce model evaluations \citep{song2020denoising,lu2022dpm}. Cache controllers retain pretrained generator weights and reuse internal computation. Static routing uses fixed schedules in DeepCache, PAB, and FORA and a learned schedule in Learning-to-Cache \citep{ma2024deepcache,zhao2025real,selvaraju2024fora,ma2024learning}. TeaCache, AdaCache, and EasyCache adapt reuse to runtime changes \citep{liu2025timestep,kahatapitiya2025adaptive,zhou2025easycache}. Recent DiT controllers allocate reuse across blocks, tokens, or trajectories \citep{chen2024deltadittrainingfreeaccelerationmethod,zou2024accelerating,zou2024DuCa,chu2025omnicache,qiu2025accelerating}. FasterCache reuses residuals, TaylorSeer forecasts features, and FlowCache supports autoregressive video \citep{lyu2025fastercache,liu2025reusing,ma2026flowcachingautoregressivevideo}. MagCache, DiCache, and SeaCache exploit magnitude, reconstruction, and spectral signals \citep{ma2026magcache,bu2025dicache,chung2026seacache}. DriveCache uses planned ego motion before denoising and uses runtime drift to veto unsupported reuse decisions.

\subsection{Autonomous Driving Video Generation}

Driving systems study occupancy, sensor fusion, planning, safety-critical generation, and scene representations \citep{zheng2024occworld,chitta2022transfuser,zheng2024genad,xing2025goalflow,xie2024advdiffuser,song2025insightdrive,duan2024maskfuser}. Surround-view studies characterize cross-view depth and spatial reasoning \citep{guo2025AdjacentView,guo2025rovr,guo2025surds}. Driving video generators forecast observations from histories, maps, layouts, and ego trajectories \citep{hu2023gaia,wang2024drivedreamer,gao2024vista,gao2024magicdrive,wen2024panacea,zhao2024drive,huang2024subjectdrive}. Multiview reconstruction models target geometric consistency and long-horizon control \citep{wang2024driving,lu2023wovogen,gao2025magicdrive,ni2025maskgwm,wu2025drivescape}, while physical-AI platforms build world foundation models \citep{nvidia2025cosmos}. Autoregressive diffusion links video to trajectories \citep{zhang2025epona,zhou2026drivinggen}; benchmarks evaluate reactive simulation and deployment robustness \citep{zhang2026reactsim,zhang2026bench2driverobust}. NuScenes provides synchronized cameras and ego-motion records \citep{caesar2020nuscenes}. DriveCache uses planned ego motion to estimate viewpoint change and allocate pre-denoising computation.

\section{Methodology}

DriveCache formulates caching as a causal decision made before each backbone evaluation. Consider a frozen video diffusion model with \(K\) denoising steps. At step \(k\), Equation~\eqref{eq:backbone-interface} separates the expensive reusable backbone from the inexpensive output interface:

\begin{equation}
r_k=F_k(u_k), \qquad \epsilon_k=H_k(x_k,r_k),
\label{eq:backbone-interface}
\end{equation}

where \(x_k\) is the current latent, \(u_k\) is the backbone input, and \(r_k\) is the cached backbone output. A full decision evaluates \(F_k\); a reuse decision replaces \(r_k\) with its most recently computed value. The first step always uses full computation. The controller observes \(u_k\) before executing \(F_k\), so it can make the veto decision without first executing the backbone call.

\begin{figure*}[t]
\centering
\includegraphics[width=0.99\textwidth]{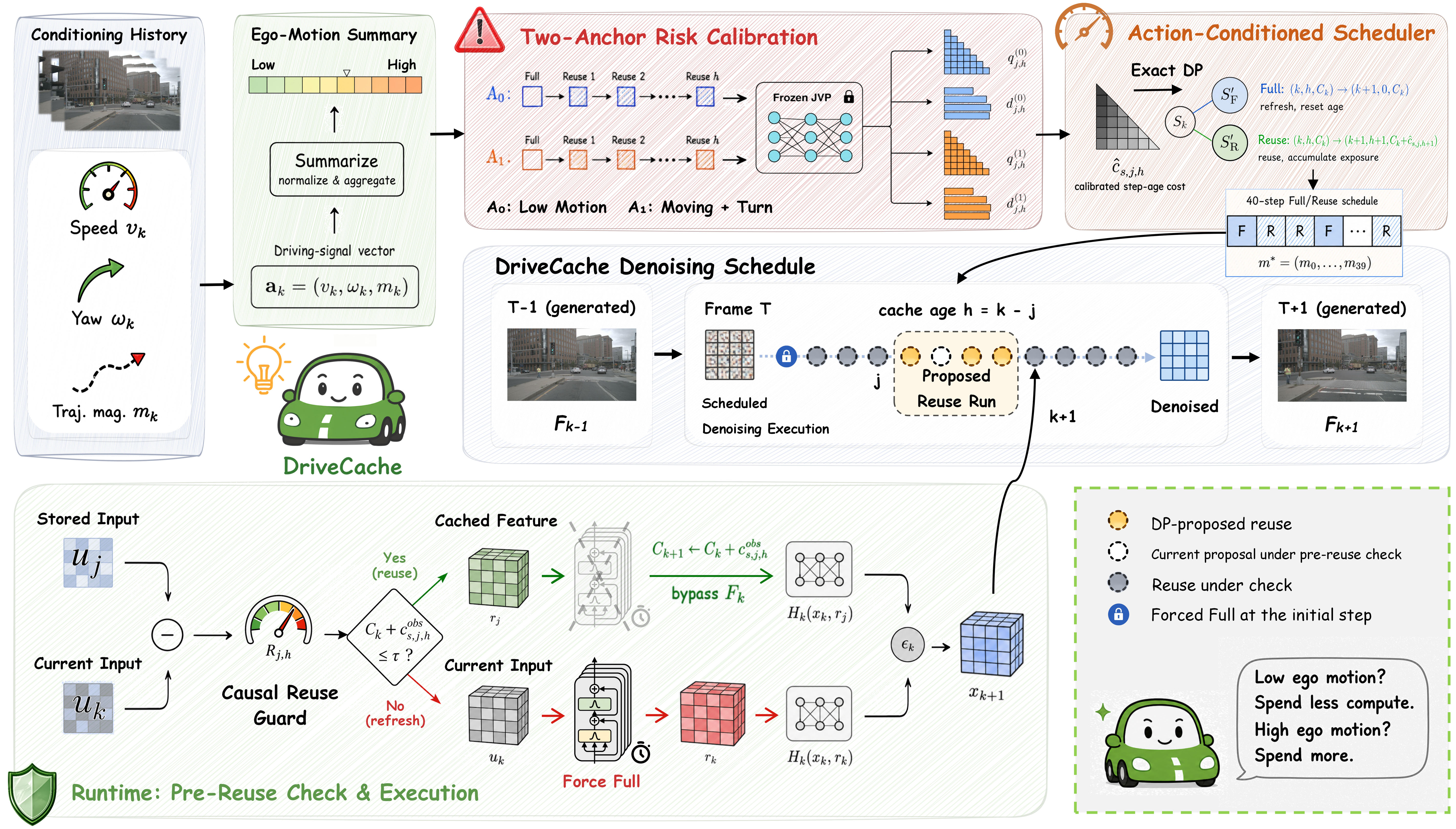}
\caption{DriveCache combines planned ego motion, two-anchor run calibration, exact DP, and a causal reuse guard.}
\label{fig:method}
\end{figure*}

DriveCache has four stages (Figure~\ref{fig:method}). Two ego-motion anchors measure joint run responses; planned ego motion interpolates them for the current scene; exact dynamic programming jointly selects reuse quantity and placement; and a pre-reuse drift check can refresh and replan the unexecuted suffix while preserving the executed prefix.

The ablation study in Table~\ref{tab:core-ablation} supports planned ego motion as a cache prior. Cache tolerance varies because planned ego motion changes the rendered viewpoint.

\begin{figure}[t]
\centering
\includegraphics[width=\columnwidth]{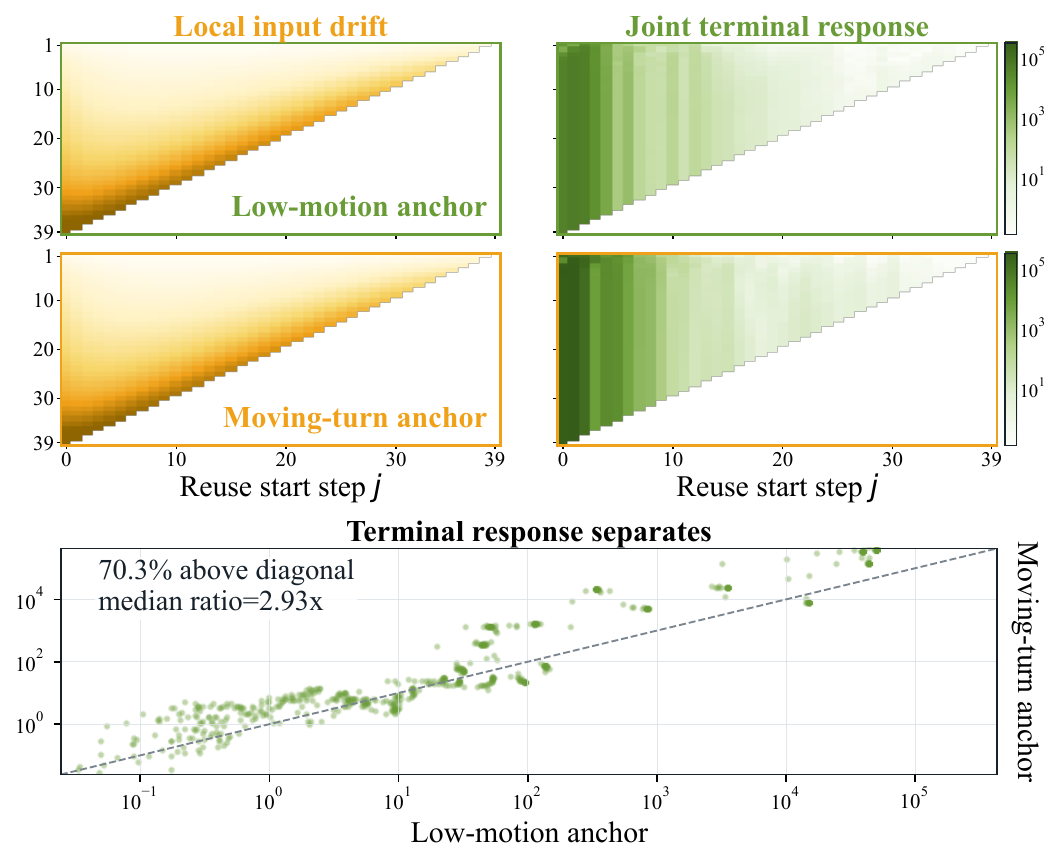}
\caption{Planned ego motion changes joint run responses even when local input drift remains nearly unchanged.}
\label{fig:motivation}
\end{figure}

Scene-agnostic scheduling, shuffled trajectories, and translation-matched turns separate planned-motion allocation from dataset correlation and fixed step preference. Figure~\ref{fig:motivation} shows that planned ego motion separates run tolerance while local input drift remains nearly unchanged. Planned ego motion remains a scene coordinate. Denoising position and cache age determine placement cost, while support checks and the causal veto handle departures from the calibrated regime.

The search enforces the mandatory first evaluation, maximum cache age, backbone boundaries, and native no-cache steps. It maximizes skipped evaluations under one cumulative response budget, with deterministic lower-cost tie breaking. Planned ego motion changes allocation before denoising, run costs place reuse jointly, and a veto corrects the next decision before a cached backbone output changes the latent. The executed prefix is never altered.

\subsection{Planned ego motion calibrates run costs}

Let a scene provide planned ego poses \((p_t,\psi_t)\) over the generated horizon. Equation~\eqref{eq:motion-statistics} summarizes total translation and accumulated rotation:

\begin{equation}
L_s=\sum_t\lVert p_t-p_{t-1}\rVert_2, \qquad
\Theta_s=\sum_t\left|\operatorname{wrap}(\psi_t-\psi_{t-1})\right|.
\label{eq:motion-statistics}
\end{equation}

The statistics remain separate because equal travel distance can induce different cached-output changes under straight and turning motion.

We z-score \(L\) and \(\Theta\) using their calibration means and standard deviations. The low-motion anchor \((L_0,\Theta_0)\) minimizes the sum of these standardized values. Among clips whose raw \(\Theta\) exceeds its calibration median and whose raw \(L>L_0\) and \(\Theta>\Theta_0\), the moving-turn anchor \((L_1,\Theta_1)\) maximizes the product of the standardized values. Clip index breaks ties, and an empty candidate set triggers full-inference fallback. Equation~\eqref{eq:motion-normalization} expresses a scene relative to this support:

\begin{equation}
z_s^L=\frac{L_s-L_0}{\max(L_1-L_0,\varepsilon)},\qquad
z_s^\Theta=\frac{\Theta_s-\Theta_0}{\max(\Theta_1-\Theta_0,\varepsilon)}.
\label{eq:motion-normalization}
\end{equation}

The anchor pair maps low motion toward the origin and the moving turn toward the upper corner without fitting scene-specific coefficients. A scene lies inside the calibration support only when \((z_s^L,z_s^\Theta)\in[0,1]^2\); otherwise DriveCache uses full inference. Equation~\eqref{eq:motion-intensity} then reduces the supported pair to one interpolation coordinate:

\begin{equation}
D_s=\frac{1}{\sqrt 2}\sqrt{(z_s^L)^2+(z_s^\Theta)^2}.
\label{eq:motion-intensity}
\end{equation}

The coordinatewise gate prevents radial compression from hiding an unsupported dimension.

The anchor traces also define how DriveCache scores a consecutive reuse run. For each anchor \(i\in\{0,1\}\), one full denoising trace stores \((x_k^{(i)},u_k^{(i)},r_k^{(i)})\). A candidate run starts after a full refresh at step \(j\) and reuses \(r_j^{(i)}\) for \(h\) steps. Equation~\eqref{eq:anchor-input-drift} measures how far the current backbone input has moved from the refresh input at the \(\ell\)-th reuse:

\begin{equation}
d_{j,\ell}^{(i)}=\frac{\lVert u_{j+\ell}^{(i)}-u_j^{(i)}\rVert_2}{\lVert u_{j+\ell}^{(i)}\rVert_2+\varepsilon}.
\label{eq:anchor-input-drift}
\end{equation}

Input drift is available before the expensive backbone call and is therefore suitable for the runtime veto. It does not by itself measure the denoising error caused by substituting a cached backbone output. Equation~\eqref{eq:local-output-perturbation} defines that signed output perturbation against the full trace:

\begin{equation}
e_{j,\ell}^{(i)}=H_{j+\ell}(x_{j+\ell}^{(i)},r_j^{(i)})-H_{j+\ell}(x_{j+\ell}^{(i)},r_{j+\ell}^{(i)}).
\label{eq:local-output-perturbation}
\end{equation}

Frozen-sampler Jacobian-vector products propagate the whole run to the terminal latent. Let \(P_t^{(i)}\) be the Jacobian of terminal latent \(x_K\) with respect to denoiser output \(\epsilon_t\), evaluated along anchor \(i\)'s frozen full trace. Its product with \(e_t^{(i)}\) gives the first-order terminal perturbation. Equation~\eqref{eq:joint-run-response} combines these signed perturbations into joint run response \(q_{j,h}^{(i)}\):

\begin{equation}
G_{j,h}^{(i)}=\sum_{\ell=1}^{h}P_{j+\ell}^{(i)}e_{j,\ell}^{(i)},\qquad
q_{j,h}^{(i)}=\frac{\lVert G_{j,h}^{(i)}\rVert_2}{\lVert x_K^{(i)}\rVert_2+\varepsilon}.
\label{eq:joint-run-response}
\end{equation}

Normalizing by terminal-latent energy makes responses comparable across steps without a learned evaluator. Replay isolates cache perturbation, while the signed vector preserves cross-step cancellation and amplification.

For scene \(s\), Equation~\eqref{eq:scene-response-interpolation} interpolates only the nonnegative planned-motion-dependent difference:

\begin{equation}
\widetilde q_{s,j,h}=q_{j,h}^{(0)}+D_s\max\!\left(q_{j,h}^{(1)}-q_{j,h}^{(0)},0\right).
\label{eq:scene-response-interpolation}
\end{equation}

This one-sided interpolation preserves the low-motion response when the moving anchor is easier and enforces a monotone planned-motion risk relation. Equation~\eqref{eq:monotone-run-cost} converts joint run response \(\widetilde q\) into monotone envelope \(Q\) and incremental run cost \(\widehat c\):

\begin{equation}
\begin{aligned}
Q_{s,j,h} &= \max_{1\leq \ell\leq h}\widetilde q_{s,j,\ell},\\
\widehat c_{s,j,h} &= Q_{s,j,h}-Q_{s,j,h-1},\qquad Q_{s,j,0}=0.
\end{aligned}
\label{eq:monotone-run-cost}
\end{equation}

Planned ego motion changes how much reuse a scene can tolerate, while \((j,h)\) captures denoising position and cache age. We obtain drift threshold \(\bar d_{s,j,h}\) with the same one-sided anchor interpolation as Equation~\eqref{eq:scene-response-interpolation}, replacing \(q\) with \(d\).

The envelope keeps longer-run increments nonnegative; refresh resets age but not response already propagated into the latent.

\subsection{Exact scheduling supports causal correction}

DriveCache converts the run costs into a legal schedule under response budget \(\tau\). For target reuse fraction \(\rho_{\mathrm{tar}}\), calibration chooses the smallest table-cost threshold whose median schedule reaches \(\lceil\rho_{\mathrm{tar}}(K-1)\rceil\) reuses; neither PSNR nor evaluation clips enter this choice. Since legal reuses skip the same interface, maximizing their count minimizes backbone evaluations. Let \(J_k(n,h)\) be minimum cumulative response after \(k\) steps, \(n\) reuses, and cache age \(h\). We set \(J_1(0,0)=0\) and all other states to \(+\infty\), with \(0\leq n\leq k-1\) and \(0\leq h\leq H_{\max}\). Here, \(\mathcal{L}(k,h)=1\) marks reuse that satisfies the model-specific legality constraints:

\begin{equation}
\begin{aligned}
J_{k+1}(n,0) &= \min_h J_k(n,h),\\
J_{k+1}(n+1,h+1) &= \min\!\bigl\{J_{k+1}(n+1,h+1),\\
&\hspace{18mm}J_k(n,h)+\widehat c_{s,k-h,h+1}\bigr\},\\
&\hspace{-12mm}\mathcal{L}(k,h+1)=1.
\end{aligned}
\label{eq:dp-transitions}
\end{equation}

A full evaluation pays no new reuse cost and resets cache age. A reuse extends the run whose last full evaluation occurred at \(j=k-h\) and adds the calibrated increment only when the next age is legal. Illegal transitions receive \(+\infty\). Equation~\eqref{eq:optimal-reuse-count} selects the maximum feasible reuse count:

\begin{equation}
B_s^*=\max\left\{n:\min_h J_K(n,h)\leq\tau\right\}.
\label{eq:optimal-reuse-count}
\end{equation}

Backtracking recovers the minimum-response schedule at \(B_s^*\). The recurrence exactly optimizes the calibrated response objective in \(O(K^3)\) time and \(O(K^2)\) rolling memory.

The schedule remains causal during inference. We initialize accumulated charge as \(C_1=0\). Before the \(h\)-th reuse in a run whose last full evaluation occurred at \(j\), DriveCache measures the observed input drift in Equation~\eqref{eq:observed-input-drift}:

\begin{equation}
d_{j,h}^{\mathrm{obs}}=\frac{\lVert u_{j+h}-u_j\rVert_2}{\lVert u_{j+h}\rVert_2+\varepsilon}.
\label{eq:observed-input-drift}
\end{equation}

It accepts reuse only when \(d_{j,h}^{\mathrm{obs}}\leq\bar d_{s,j,h}+\varepsilon\) and accumulated charge satisfies \(C_k+\widehat c_{s,j,h}\leq\tau\). Acceptance sets \(C_{k+1}=C_k+\widehat c_{s,j,h}\). A failed check evaluates the backbone, keeps \(C_{k+1}=C_k\), and refreshes the cache. Replanning starts from age zero after this forced evaluation and maximizes additional suffix reuse under budget \(\tau-C_k\); the executed prefix remains fixed. Calibration uses no optimizer or parameter update, and inference leaves generator weights and sampling steps unchanged at runtime.

DriveCache applies model-specific legality to the same recurrence. Here, \(u_k\), \(r_k\), and \(H_k\) are the reusable-region input, cached backbone output, and remaining sampler interface. Wan2.2 separates guidance branches; A14B also forbids cross-expert reuse. Epona caches only within each 100-step visual denoising call, never across autoregressive segments. Cache tensors, legal steps, maximum age, norm axes, and ties are frozen before evaluation.

Two full anchor traces support all candidate runs. Building the \((j,h)\) table costs \(O(K^2)\) substitutions and \(O(K^3)\) unbatched Jacobian-vector products once per configuration, without evaluation clips. On one H20, calibration takes 4 minutes for 5B, 9 minutes for A14B, and 16 minutes for Epona, with at most 6.1 GB additional memory and response tables below 2 MB. Initial DP solves take 0.8, 1.4, and 9.2 ms, respectively; drift checks and suffix replanning add less than 0.7\% to end-to-end latency.

DriveCache updates no weights and trains no predictor; one model-level calibration applies across clips.
\section{Experiments}

\subsection{Experimental setup}

\textbf{Models and protocol.} We evaluate Wan2.2 5B and expert-routed A14B on 222 DrivingGen scenes with ten fixed seeds (2,220 samples), and Epona on all 150 nuScenes validation scenes \citep{wan2025wan,zhou2026drivinggen,zhang2025epona}. Wan uses 40 denoising steps, while Epona uses 100 steps in each visual denoising call. Calibration uses 32 disjoint scenes; before evaluation, we freeze the reuse target and \(\tau\) at 50\% for Wan and 60\% for Epona.

\begin{figure*}[!t]
\centering
\includegraphics[width=\textwidth]{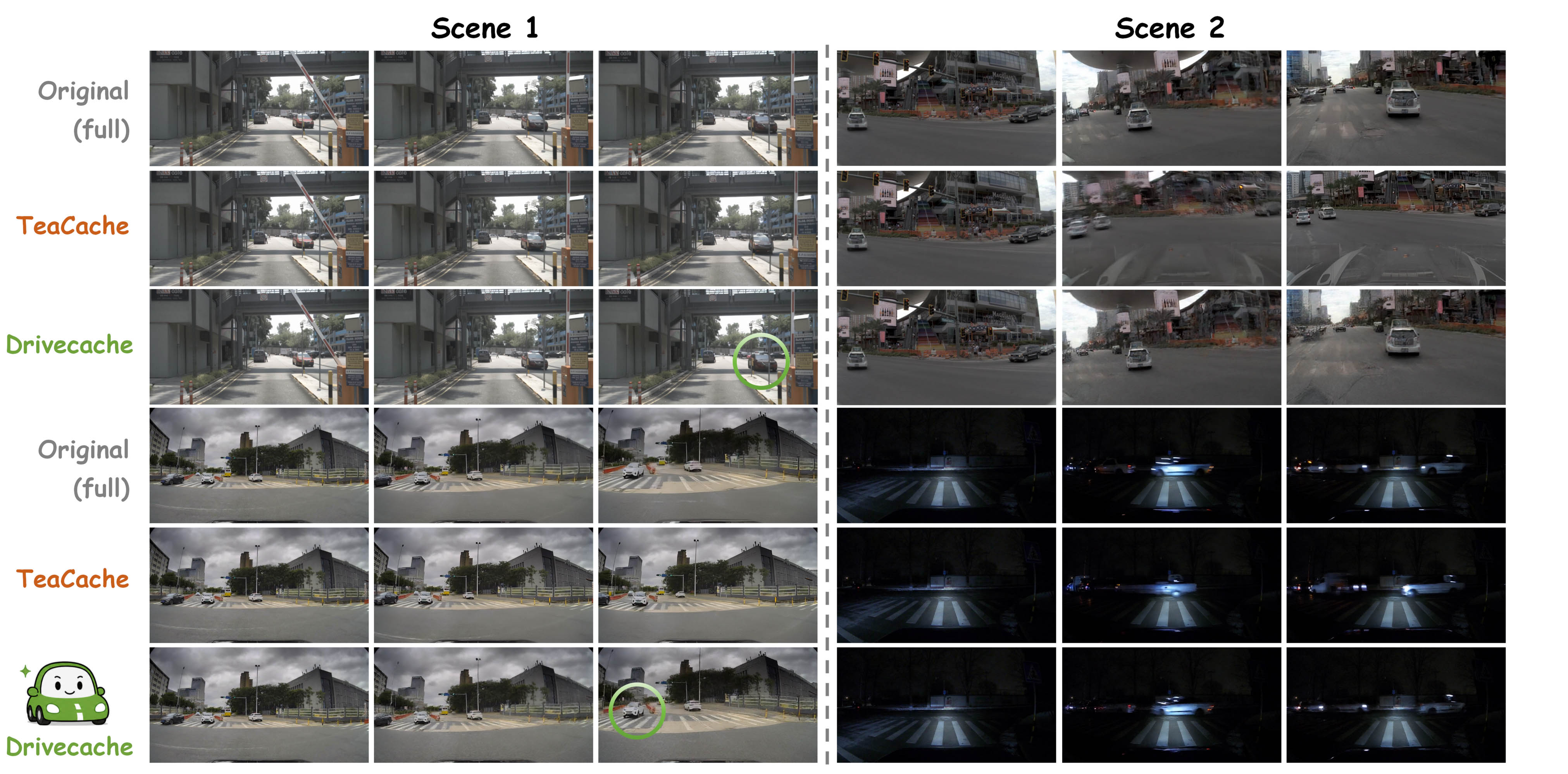}
\caption{Qualitative comparison of full inference, TeaCache, and DriveCache.}
\label{fig:qualitative}
\vspace{10pt}
\centering
\small
\renewcommand{\arraystretch}{1.08}
\setlength{\tabcolsep}{2.2pt}
\begin{tabularx}{\textwidth}{@{}l*{10}{Y}@{}}
\toprule
Backbone / Method & Lat. (s) $\downarrow$ & Spd. $\uparrow$ & PSNR $\uparrow$ & SSIM $\uparrow$ & LPIPS $\downarrow$ & V-Q $\uparrow$ & V-S $\uparrow$ & Traj. $\uparrow$ & ADE $\downarrow$ & DTW $\downarrow$ \\
\midrule
\multicolumn{11}{l}{\textbf{Wan2.2 5B}} \\
Fixed interval & 10.17 & 1.77$\times$ & 28.417 & 0.852 & 0.095 & 0.774 & \textbf{0.179} & 0.281 & 1.881 & 15.517 \\
PAB & 17.53 & 1.03$\times$ & 33.344 & 0.953 & 0.042 & 0.774 & \textbf{0.179} & 0.262 & 2.191 & 18.454 \\
TeaCache & 10.76 & 1.68$\times$ & 32.874 & 0.939 & 0.042 & 0.776 & 0.177 & 0.274 & 1.796 & 15.055 \\
EasyCache & 10.67 & 1.69$\times$ & 33.923 & 0.959 & 0.041 & 0.777 & 0.178 & 0.269 & 1.925 & 16.173 \\
MagCache & 10.73 & 1.68$\times$ & 33.945 & 0.949 & \textbf{0.038} & 0.773 & 0.177 & 0.261 & 2.712 & 23.156 \\
DiCache & 12.15 & 1.48$\times$ & 33.278 & 0.949 & \textbf{0.038} & \textbf{0.781} & \textbf{0.179} & 0.262 & 1.919 & 16.167 \\
TaylorSeer & 10.19 & 1.77$\times$ & 27.970 & 0.847 & 0.094 & 0.770 & 0.178 & 0.262 & 2.306 & 19.567 \\
SeaCache & 10.07 & 1.79$\times$ & 32.860 & 0.932 & 0.048 & 0.769 & 0.176 & 0.275 & 1.929 & 16.004 \\
\rowcolor{gray!12}\textbf{DriveCache} & \textbf{9.79} & \textbf{1.84$\times$} & \textbf{34.744} & \textbf{0.961} & 0.041 & 0.777 & \textbf{0.179} & \textbf{0.294} & \textbf{1.771} & \textbf{14.768} \\
\multicolumn{11}{l}{\textbf{Wan2.2 A14B}} \\
Fixed interval & 86.64 & 1.89$\times$ & 25.385 & 0.802 & 0.159 & 0.777 & 0.178 & 0.336 & 2.610 & 20.836 \\
PAB & 156.45 & 1.04$\times$ & 31.422 & 0.931 & 0.079 & 0.776 & 0.181 & 0.354 & 1.980 & 15.475 \\
TeaCache & \textbf{82.87} & \textbf{1.97$\times$} & 30.742 & 0.913 & 0.064 & 0.779 & 0.179 & 0.324 & 2.009 & 15.694 \\
EasyCache & 89.44 & 1.83$\times$ & 31.461 & 0.922 & 0.070 & \textbf{0.781} & 0.181 & 0.349 & 2.020 & 15.576 \\
MagCache & 86.83 & 1.88$\times$ & 31.987 & 0.937 & \textbf{0.049} & 0.777 & 0.180 & 0.346 & 2.138 & 16.858 \\
DiCache & 87.00 & 1.88$\times$ & 29.927 & 0.897 & 0.096 & 0.775 & 0.180 & 0.323 & 1.951 & 15.011 \\
TaylorSeer & 87.11 & 1.88$\times$ & 26.339 & 0.817 & 0.143 & 0.780 & 0.180 & 0.357 & 2.320 & 18.439 \\
SeaCache & 83.09 & \textbf{1.97$\times$} & 31.502 & 0.923 & 0.060 & 0.779 & 0.179 & 0.329 & 1.948 & 14.881 \\
\rowcolor{gray!12}\textbf{DriveCache} & 83.10 & \textbf{1.97$\times$} & \textbf{32.778} & \textbf{0.938} & 0.060 & \textbf{0.781} & \textbf{0.182} & \textbf{0.370} & \textbf{1.809} & \textbf{13.750} \\
\bottomrule
\end{tabularx}
\captionof{table}{Wan2.2 5B and A14B results at matched skipped-backbone counts.}
\label{tab:1}
\vspace{0pt}
\end{figure*}

\textbf{Comparisons and metrics.} Tables~\ref{tab:1}--\ref{tab:2} cover fixed, adaptive, forecasting, spectral, and reduced-step baselines \citep{ma2024deepcache,zhao2025real,lyu2025fastercache,liu2025timestep,kahatapitiya2025adaptive,zhou2025easycache,ma2026magcache,bu2025dicache,liu2025reusing,chung2026seacache,ma2026flowcachingautoregressivevideo}. We match Wan caches within one skipped backbone call per clip and compare reduced-step sampling at matched latency. Epona methods share hardware while retaining their native reusable units. We report peak signal-to-noise ratio (PSNR), structural similarity (SSIM), learned perceptual image patch similarity (LPIPS), trajectory quality (Traj.), average displacement error (ADE), dynamic time warping (DTW), and VBench metrics \citep{zhang2018unreasonable,huang2024vbench}. V-Q and V-S denote VBench quality and semantic scores. Higher PSNR, SSIM, Traj., and VBench scores indicate better quality; lower LPIPS, ADE, and DTW indicate smaller errors.

\textbf{Compute matching.} Calibration selects each Wan cache baseline's setting nearest the shared skipped-backbone target. Evaluation freezes that setting and verifies the one-call tolerance from realized backbone counts. Cache methods retain all 40 sampler steps, while the reduced-step baseline changes the sampling trajectory.

\textbf{Measurement.} On one H20 at batch size one, latency averages ten synchronized runs after three warm-ups and includes controller overhead. Fidelity uses matched-seed full outputs. DrivingGen evaluates trajectory quality, ADE, and DTW against ground-truth ego trajectories, while VBench uses its native references. We average seeds per scene and report paired scene-level 95\% bootstrap intervals.

\textbf{Run integrity.} Each sample records its configuration hash, realized backbone calls, and metric outputs. We retain complete 2,220-sample DrivingGen runs and complete 150-scene Epona validation runs for comparison.

\begin{table*}[!t]
\centering
\small
\renewcommand{\arraystretch}{0.88}
\setlength{\tabcolsep}{4pt}
\begin{tabularx}{\textwidth}{@{}l*{7}{Y}@{}}
\toprule
Method & Latency $\downarrow$ & Speedup $\uparrow$ & PSNR $\uparrow$ & SSIM $\uparrow$ & LPIPS $\downarrow$ & Stationary PSNR $\uparrow$ & Turn PSNR $\uparrow$ \\
\midrule
Reduced steps & 46.34 & 1.61$\times$ & 26.322 & 0.729 & 0.144 & 31.165 & 23.855 \\
DeepCache & 44.27 & 1.68$\times$ & 26.358 & 0.730 & 0.142 & 31.192 & 23.898 \\
FasterCache & 46.46 & 1.60$\times$ & 26.312 & 0.729 & 0.144 & 31.155 & 23.847 \\
TeaCache & 47.28 & 1.58$\times$ & 26.152 & 0.726 & 0.146 & 30.800 & 23.744 \\
AdaCache & 44.07 & 1.69$\times$ & 26.328 & 0.729 & 0.144 & 31.154 & 23.870 \\
FlowCache & 51.57 & 1.44$\times$ & 26.322 & 0.729 & 0.144 & 31.123 & 23.885 \\
\rowcolor{gray!12}\textbf{DriveCache} & \textbf{39.55} & \textbf{1.89$\times$} & \textbf{26.394} & \textbf{0.734} & \textbf{0.140} & \textbf{31.479} & \textbf{23.910} \\
\bottomrule
\end{tabularx}
\caption{DriveCache results on Epona. The method reaches \(1.89\times\) speedup at video quality comparable to cache baselines.}
\label{tab:2}
\end{table*}

\begin{figure}[!t]
\centering
\includegraphics[width=\linewidth]{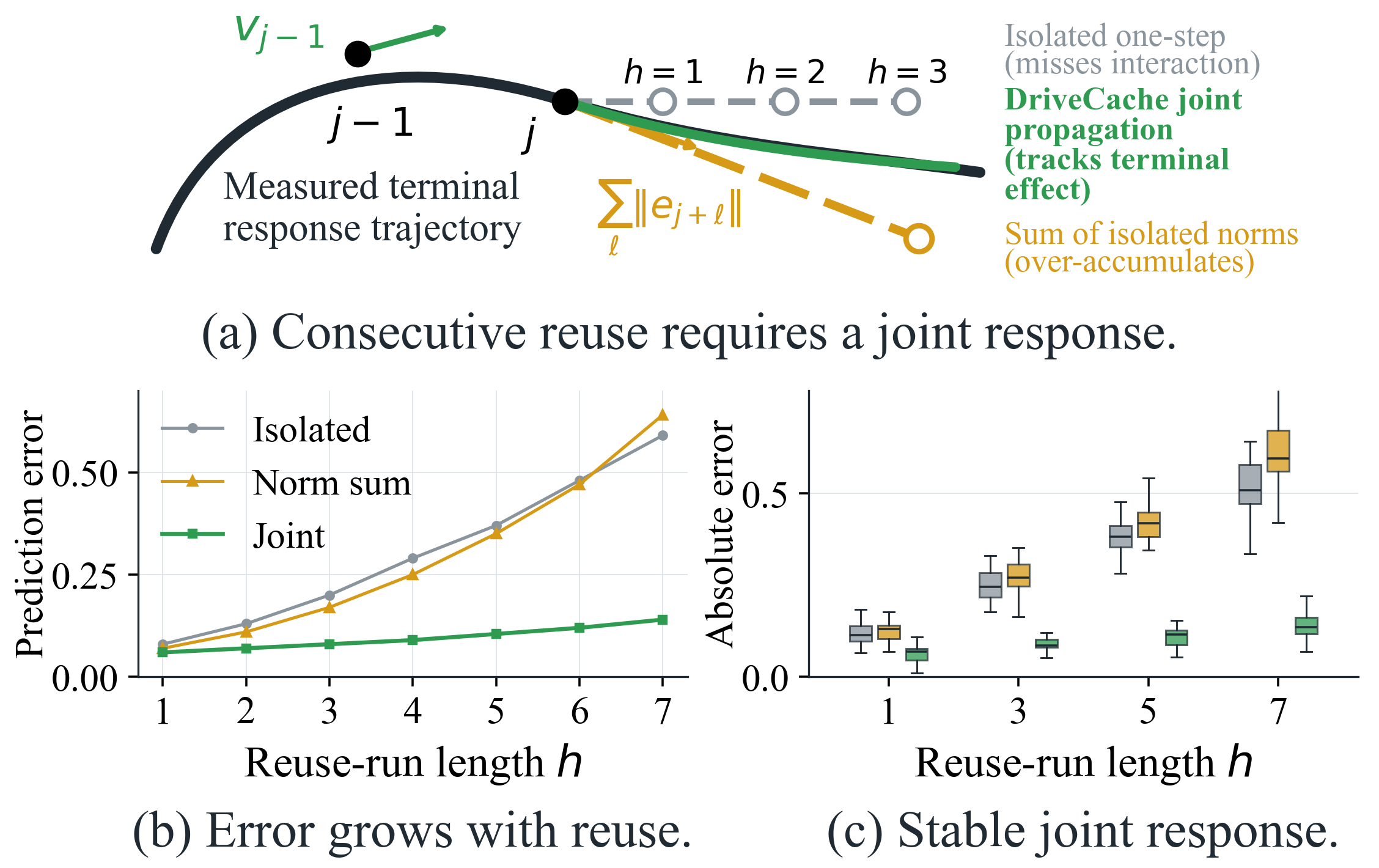}
\captionsetup{skip=1pt}
\caption{Joint propagation captures reuse-run interactions.}
\label{fig:joint-run-response}
\vspace{2pt}
\includegraphics[width=\linewidth]{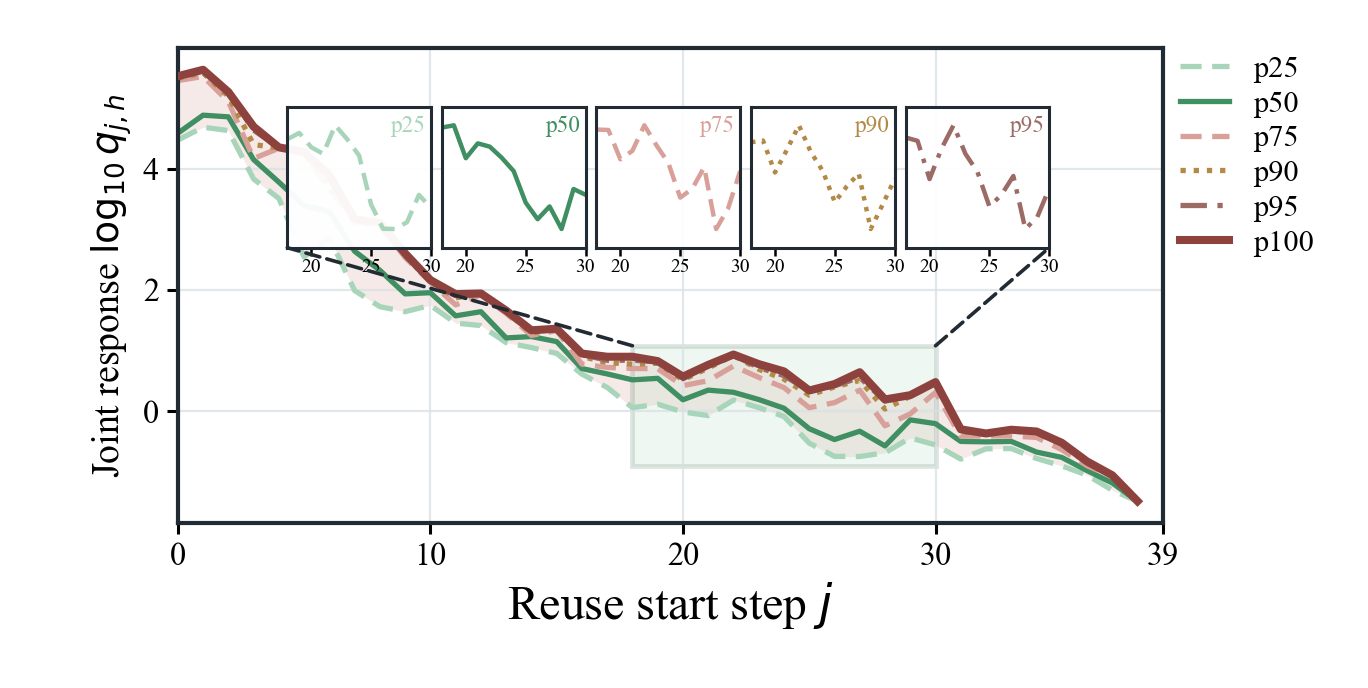}
\captionsetup{skip=1pt}
\captionof{figure}{Joint responses vary with reuse start step and cache age, guiding reuse placement across denoising steps.}
\label{fig:joint-response-percentiles}
\end{figure}

\begin{table}[!t]
\centering
\small
\renewcommand{\arraystretch}{0.88}
\captionsetup{skip=3pt}
\setlength{\tabcolsep}{2.6pt}
\begin{tabular}{lccc}
\toprule
Variant & PSNR $\uparrow$ & LPIPS $\downarrow$ & Traj. $\uparrow$ \\
\midrule
\rowcolor{gray!12}\textbf{DriveCache} & \textbf{34.744} & \textbf{0.041} & \textbf{0.294} \\
No motion prior & 34.201 & 0.046 & 0.274 \\
Shuffled trajectory & 34.286 & 0.045 & 0.275 \\
Translation only & 34.431 & 0.043 & 0.278 \\
Rotation only & 34.365 & 0.044 & 0.277 \\
No DP (fixed placement) & 33.982 & 0.050 & 0.272 \\
No drift veto & 34.512 & 0.043 & 0.279 \\
\bottomrule
\end{tabular}
\captionof{table}{Wan2.2 5B ablations at 21/40 realized reuse isolate motion allocation, step placement, and runtime correction.}
\label{tab:core-ablation}
\vspace{10pt}
\centering
\small
\renewcommand{\arraystretch}{0.88}
\captionsetup{skip=3pt}
\setlength{\tabcolsep}{2.2pt}
\begin{tabular}{lcccc}
\toprule
High/low reuses & Lat. $\downarrow$ & PSNR $\uparrow$ & SSIM $\uparrow$ & LPIPS $\downarrow$ \\
\midrule
\textbf{0/20} & \textbf{83.10} & \textbf{32.778} & \textbf{0.938} & \textbf{0.060} \\
2/18 & 83.02 & 30.747 & 0.914 & 0.073 \\
3/17 & 83.34 & 29.936 & 0.900 & 0.081 \\
\bottomrule
\end{tabular}
\captionof{table}{A14B expert routing at fixed compute. High/low denotes reuse counts in the high- and low-noise experts.}
\label{tab:a14b-routing}
\end{table}
\vspace{-10pt}

\subsection{DriveCache improves quality at matched speed}

At the frozen operating points in Table~\ref{tab:1}, DriveCache reaches \(1.84\times\) speedup and 34.744 dB PSNR on 5B, and \(1.97\times\) and 32.778 dB on A14B. Viewed jointly, it is fastest on 5B and Epona and tied for the highest speedup on A14B, while retaining the best values in most reported quality and driving dimensions. Figure~\ref{fig:teaser} shows DriveCache's quality gains across the evaluated driving motions in every motion group.

\textbf{Fidelity and downstream behavior.} At comparable speed, DriveCache improves over SeaCache by 1.884 dB PSNR (95\% CI: 1.61--2.15) on 5B and by 1.276 dB (1.04--1.51) on A14B; trajectory quality and DTW also improve. Trajectory metrics show ego-motion consistency; Figure~\ref{fig:qualitative} and VBench show temporal consistency and stable scene semantics.

\textbf{Motion adaptation and robustness.} The anchors cover 96.4\% of scenes; stationary clips average 22.4 reuses versus 18.7 for turns, confirming motion-conditioned allocation. The motion prior uses planned ego motion to set the schedule for every supported scene before the first denoising step. The~drift check evaluates generated states before each planned reuse during inference. Under 1.0 m/5 degree pose noise, DriveCache retains \(1.57\times\) speedup with a 0.213 dB loss, showing that causal correction complements the motion prior.

\textbf{Transfer to Epona.} Across 150 Epona scenes, DriveCache reaches \(1.89\times\) speedup and 0.140 LPIPS, versus DeepCache's \(1.68\times\) at comparable aggregate and turning fidelity (Table~\ref{tab:2}). Larger stationary gains reflect more reuse on tolerant scenes and conservative behavior on turns. The Epona results extend the latency--fidelity gains to a separate scene collection.

\subsection{Ablations validate scheduling choices}

\textbf{Planned motion controls scene-level reuse.} Mean speed correlates with cache error, while removing or shuffling motion degrades fixed-reuse quality. Shuffling preserves the trajectory distribution but breaks scene correspondence, isolating action--scene alignment; translation and rotation both contribute to cache allocation.

\textbf{Joint responses control placement.} Figures~\ref{fig:joint-run-response}--\ref{fig:joint-response-percentiles} show that joint propagation captures start-step and cache-age effects missed by isolated sums. Fixed placement causes the largest loss, while the drift veto guards against schedule drift.

\textbf{Calibration and routing remain stable.} Predicted response tracks realized degradation, and two to eight anchors change PSNR by only 0.062 dB. Moving reuse into the high-noise A14B expert sharply reduces quality (Table~\ref{tab:a14b-routing}), so expert boundaries matter more than richer interpolation at fixed compute. The core ablations in Table~\ref{tab:core-ablation} hold the pretrained generator fixed while isolating scene allocation, step placement, and runtime correction. Together, the anchor and routing results support sparse calibration and explicit expert boundaries in the controller.
\vspace{4pt}
\section{Conclusion}
Across the evaluated driving motions, planned ego translation and rotation predict scene-level cache tolerance before denoising. We introduce DriveCache, a training-free controller that allocates reuse across scenes and places it across denoising steps with exact dynamic programming. Two anchor traces calibrate joint run responses, and a causal drift check refreshes the cache and replans the unexecuted suffix when generation departs from calibration. On Wan2.2 5B, Wan2.2 A14B, and Epona, DriveCache improves quality--efficiency trade-offs over cache baselines while preserving generator weights and sampling steps. These results support planned ego motion as a control signal for action-aware diffusion caching.
\clearpage
{\small
\setlength{\bibsep}{0pt plus 0.2ex}
\bibliography{references}
}
\end{document}